\documentclass[conference]{IEEEtran}
\IEEEoverridecommandlockouts
\usepackage{cite}
\usepackage{amsmath,amssymb,amsfonts}
\usepackage{algorithmic}
\usepackage{graphicx}
\usepackage{textcomp}
\usepackage{xcolor}
\usepackage{makecell}
\usepackage{pifont}
\usepackage{hhline}
\usepackage{scalerel,xparse}
\usepackage{threeparttable}
\usepackage{paralist}
\usepackage{subcaption}
\usepackage{tabu}
\usepackage{color, colortbl}
\usepackage[accsupp]{axessibility}

\usepackage{float}

\usepackage{tabularx}
\usepackage{booktabs}
\usepackage{multirow}
\usepackage{array,hhline}
\usepackage{subcaption}
\usepackage{wrapfig}
\usepackage{enumitem}
\usepackage{diagbox}

\definecolor{Gray}{gray}{0.5}
\definecolor{LGray}{gray}{0.9}
\definecolor{darkblue}{RGB}{94,110,186}
\definecolor{darkGreen}{RGB}{92, 180, 110}
\definecolor{myblue}{RGB}{14, 121, 178}
\definecolor{myred}{RGB}{192, 0, 0}

\newcommand{\darkGreen}[1]{\textcolor{darkGreen}{#1}}
\newcommand{\myblue}[1]{\textcolor{myblue}{#1}}
\newcommand{\myred}[1]{\textcolor{myred}{#1}}

\def\BibTeX{{\rm B\kern-.05em{\sc i\kern-.025em b}\kern-.08em
    T\kern-.1667em\lower.7ex\hbox{E}\kern-.125emX}}

\begin{document}

\title{UCVL: A Benchmark for Crime Surveillance Video Analysis with Large Models}
\author{
  Haoran Chen\textsuperscript{1,2,3}, Dong Yi\textsuperscript{1,2,3}, Moyan Cao\textsuperscript{2}, Chensen Huang\textsuperscript{1,2,3}, Guibo Zhu\textsuperscript{1,2,3}, Jinqiao Wang\textsuperscript{1,2,3}\\
  \textsuperscript{1}Institute of Automation, Chinese Academy of Sciences\\
  \textsuperscript{2}University of Chinese Academy of Sciences\\
  \textsuperscript{3}Wuhan AI Research
}

\maketitle
\begin{abstract}
  Anomaly analysis in surveillance videos is a crucial topic in computer vision. 
  In recent years, multimodal large language models (MLLMs) have outperformed task-specific models in various domains. 
  Although MLLMs are particularly versatile, their abilities to understand anomalous concepts and details are insufficiently studied because of the outdated benchmarks of this field not providing MLLM-style QAs and efficient algorithms to assess the model's open-ended text responses. 
  To fill this gap, we propose a benchmark for crime surveillance video analysis with large models denoted as UCVL, including 1,829 videos and reorganized annotations from the UCF-Crime and UCF-Crime Annotation datasets. We design six types of questions and generate diverse QA pairs. Then we develop detailed instructions and use OpenAI's GPT-4o for accurate assessment. 
  We benchmark eight prevailing MLLMs ranging from 0.5B to 40B parameters, and the results demonstrate the reliability of this bench. 
  Moreover, we finetune LLaVA-OneVision on UCVL's training set. The improvement validates our data's high quality for video anomaly analysis.
  \footnote{The UCVL dataset, evaluation code and model checkpoints will be released after this paper is published.}
 \end{abstract}


 \section{Introduction}
 \label{sec:intro}

   Surveillance videos play an important role in security governance and crime investigation.
   However, it remains a great challenge to detect complex human activities or ground the temporal location of events from prolonged and low-quality videos.
   Existing methods are often trained on task-specific datasets\cite{UCFCrime,PDVC,BN-WVAD,MMN}, which address only one aspect of this challenge with each model, such as anomaly detection\cite{UCFCrime}, temporal grounding\cite{MMN}, or video captioning\cite{PDVC}.
   Although these task-specific models work well in their own domain, but they fall short in providing detailed descriptions and reasoning steps, which could give security officers a quick insight into sophisticated cases. 
   On the other hand, task-specific models are inconvenient for synthesizing information from various aspects.
 
\begin{figure}[t]
    \centering
    \includegraphics[width=\linewidth]{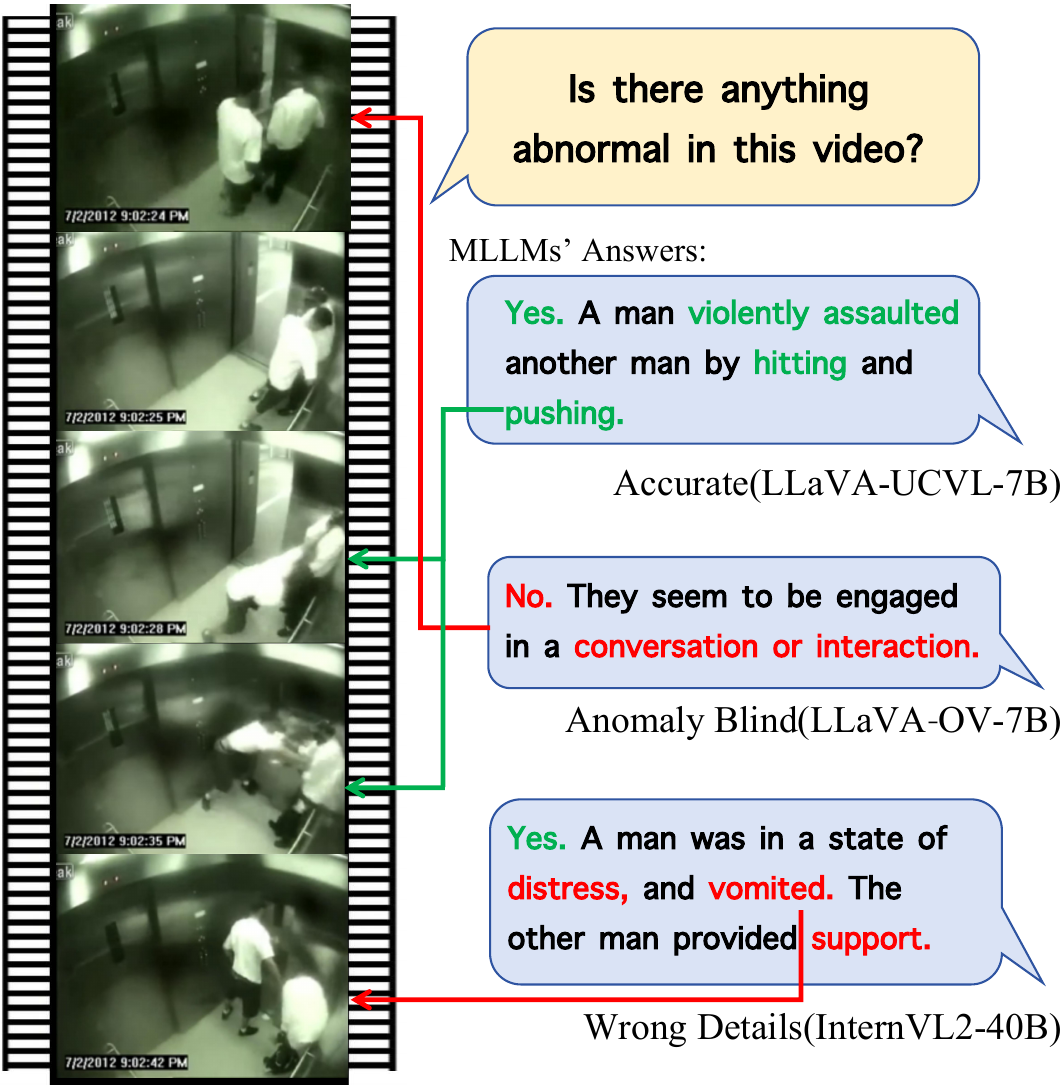}
    \caption{A comparison of different models' performance on an ''Assault'' video from UCVL. The \darkGreen{green lines} highlight the correct answers and descriptions, while the \myred{red lines} indicate wrong answers and descriptions. See more cases in Appendix A.}
    \label{fig:fig1}
\end{figure}

   In recent years, Transformer-based large models have witnessed rapid advancements\cite{qwen,llava-onevision,internvl,qwen2VL}. 
   The alignment of visual and text features has allowed the models to process multimodal data with remarkable versatility\cite{CLIP,qwen2VL}, showing promise in tackling all tasks with a single model.
   Although many MLLM benchmarks often claim to evaluate dozens of cross-modal capabilities\cite{mm-vet,MMBench,mmbench-video,mvbench,video-mme}, the ability to perceive anomalous events or concepts has not been thoroughly examined. 
   We attribute this negligence to the long-standing practice of filtering multimodal data, which often excludes content not-suitable-for-work (NSFW) and low-quality data\cite{TaiSu}.
   Therefore, MLLMs may exhibit cognitive blindness in recognizing such scenes, as shown in Fig. \ref{fig:fig1}. 
   Moreover, existing anomaly benchmarks are not qualified to quantitatively evaluate MLLM due to their task-specific output structures and evaluation metrics\cite{XD-violence,UCFCrime}.
 
   To address these challenges, we introduce UCVL, the first MLLM benchmark designed for multi-task anomaly analysis in crime surveillance video.
   To avoid redundant effort, we fully leverage the previous anomaly datasets by
   integrating the crime labels from UCF-Crime (UCC)\cite{UCFCrime} and the segment-level human descriptions from UCF-Crime Annotation (UCA)\cite{UCA}.
\begin{figure*}[htbp]
    \centering
    \includegraphics[width=1\linewidth]{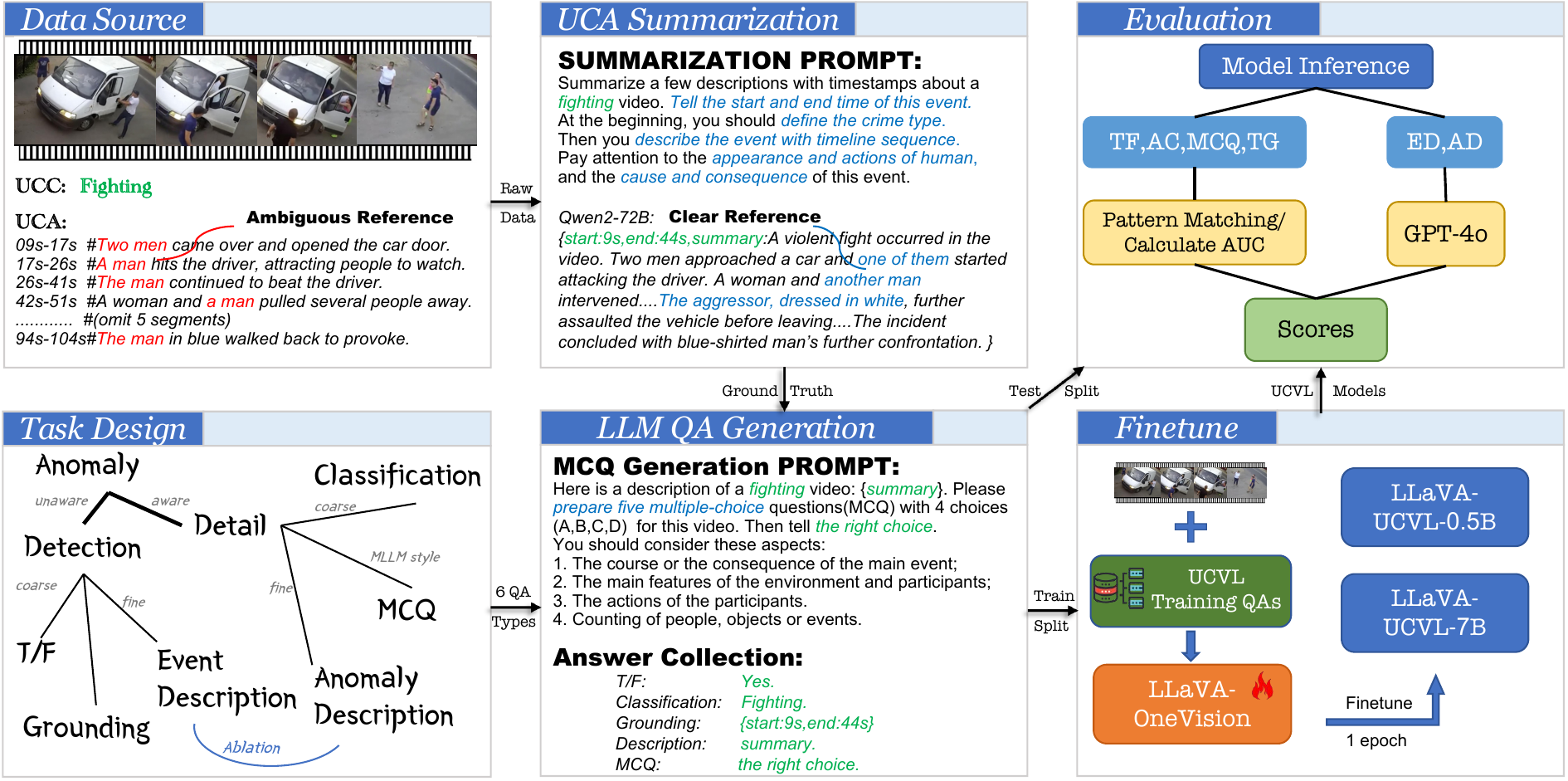}
    \caption{The pipeline of this work. We first parse data source and design task types. Then we use LLM to generate the video summary and QA pairs. Finally, we finetune two models and evaluate ten models on UCVL.}
    \label{fig:pipeline}
  \end{figure*} 
   Specifically, we design multi-task questions following the general MLLM benchmarking patterns\cite{video-mme,mvbench,mmbench-video} and instruct a powerful large language model (LLM)\cite{qwen} to generate question-answer pairs (QAs) based on the human labeled ground truth for each video.
   To evaluate the capabilities of MLLMs, we propose a rigorous scoring system that uses pattern matching for objective questions and GPT-4o\cite{gpt4o} for open-ended responses with more detailed criteria.
   Compared with UCC and UCA, UCVL highlights three main features: 
 (1) UCVL reformulates conventional crime anomaly analysis tasks into a unified QA format for MLLMs, therefore assessing MLLMs' versatility without adaptation.
 (2) UCVL evaluates MLLMs from multiple aspects with diverse QAs generated by Qwen2 LLM.
 (3) The structured prompting and questioning enable objective scoring for model's answers, 
    while detailed and strict rules make GPT-4o effective for evaluating open-ended description tasks.

   We evaluate eight prevalent MLLMs with video abilities on the test set of 300 videos and compare their performance across different aspects of anomaly analysis.
   Additionally, we finetune two of these models on the training split of 1030 videos, leading to encouraging enhancements.

   \vspace{5pt}
   In conclusion, our work has three main contributions: 
   \begin{itemize}[noitemsep,nolistsep,leftmargin=*]

     \item We propose UCVL, the first MLLM benchmark for crime surveillance video analysis, with detailed and effective evaluation criteria. 
  \item We first point out the blindness of MLLMs in perceiving anomalous concepts and actions in crime surveillance video by comprehensive evaluation.
     \item The training split of UCVL is used to finetune two models, whose significant performance improvements demonstrate the high quality of this dataset and showcase the potential of MLLMs in video anomaly analysis.
   \end{itemize}

 \section{Related Work}
 \label{sec:related_work}
 \subsection{Video Anomaly Analysis} 
   Analyzing a video's anomaly includes many aspects.
 For the detection task, BN-WVAD\cite{BN-WVAD} embeds videos with I3D\cite{I3D} and enhances the features with a Transformer. A convolutional classifier then outputs an anomaly score.
 For the captioning task, PDVC\cite{PDVC} uses C3D\cite{C3D} for feature extraction, a Transformer-based encoder-decoder for multi-task prediction, and an LSTM for caption generation.
 For the temporal grounding task, MMN\cite{MMN} employs C3D and DistillBERT\cite{DistilBERT} to embed both video and text queries into a shared space for mutual matching.
These top-performing methods highlight the integration of traditional CNNs with Transformers; however, this approach undermines the inherent generalization capabilities of Transformer architectures.
 
 Meanwhile, the different metrics and annotations across benchmarks for different tasks make it challenging to evaluate general models.
 UCC\cite{UCFCrime} includes only abnormal categories and evaluates classification performance using Area Under Curve (AUC). Furthermore, UCA\cite{UCA} segments the UCC videos and annotates each part with a sentence, allowing evaluation of temporal grounding using Intersection over Union (IoU) and captioning using BLEU\cite{BLEU}.

 \begin{table*}[ht]
  \centering
  \setlength\tabcolsep{4pt}
  \renewcommand{\arraystretch}{1.2}
  {\Large
  \resizebox{1.0\textwidth}{!}{
      \begin{tabular}{c|c|l|p{0.1\textwidth}}
      \Xhline{1.0pt}
      \textbf{Task} & \textbf{Capability} & \textbf{QA Example} & \textbf{Truth} \\
      \Xhline{1.0pt}
      \textbf{Anomaly} & Anomaly   & \textit{\myblue{Is there any abnormal event that might be related to violence, crime or danger in this video?}} &  \multirow{2}{*}{\centering Yes} \\
      \textbf{Detection}  & Perception &  \textit{\myblue{You should only answer Yes or No.} }                  & ~ \\
      \Xhline{1.0pt}
      \textbf{Anomaly}         & Coarse   & \textit{\myblue{We believe that a dangerous event occurred in this video. Identify its category.}} &  \multirow{2}{*}{Assault} \\
      \textbf{Classification}  & Analysis &  \textit{\myblue{Your Answer should only consist of the categories from \texttt{[13 crime types]}.}}                 & ~ \\
      \Xhline{1.0pt}
      \textbf{}              & Causal   & \textit{Who was the main perpetrator in the video? } (A)The police officer. (B)The man in a white T-shirt. &  \multirow{2}{*}{\centering B} \\
      \textbf{}             & Reasoning &  { (C)The woman in white clothes.  (D)The individual pointing at the suspect.}    & ~ \\
      \cline{2-4}
      \textbf{}         & Object Detail   & \textit{How was the suspect transported away?} (A) Stretcher. (B) Car. (C) Electric scooter. (D) Bike. &  A \\
      \cline{2-4}
      \textbf{Multi-Choice} & & \textit{{What action did the male police officer in a brown uniform take towards the man in the white T-shirt?}} &  \multirow{2}{*}{\centering C}\\
      \textbf{Question}                & Action & {(A) Shot him. (B) Walked away. (C) Subdued and forced him to the ground. (D) Ignored him.} &  ~     \\
      \cline{3-4}
      \textbf{(MCQ)}                & Recognition  & \textit{{What was the response of the man in the white T-shirt when additional officers tried to handcuff him?}} &  \multirow{2}{*}{\centering D}\\
      \textbf{} & & {(A) He cooperated. (B) He ran away. (C) He surrendered. (D) He showed resistance.} &  ~     \\
      \cline{2-4}
       ~         & \multirow{2}{*}{Count}  & \textit{{How many black police officers joined the arrest operation later in the video?}} &  \multirow{2}{*}{\centering B}\\
        ~ & ~ & (A) One. (B) Two. (C) Three. (D) None. & ~\\
      \Xhline{1.0pt}
       \textbf{}      & Anomaly & \textit{\myblue{The video lasts for} 105 \myblue{seconds, and} 32 \myblue{frames are uniformly sampled from it. These frames are}}  & \{"start\_ \\
       \textbf{Time} & Perception  & \textit{\myblue{located at} \texttt{[32 timestamps]}\myblue{. Detect an abnormal event of Arrest in it and locate the start time}}  &time":21,  \\ 
       \cline{2-2}
        \textbf{Grounding}  &Causal  &\textit{\myblue{and the end time of this event.}} & "end\_time"\\
         &Reasoning &\textit{\myblue{Your answer should ONLY be \{"start\_time": start\_time, "end\_time": end\_time\}.}}   &  :109\}   \\
      \Xhline{1.0pt}
      \textbf{}                 &Anomaly      & \multicolumn{2}{l}{(For AD) \textit{\myblue{We believe an event of} Arrest \myblue{occurs in this video.}} (For both) \textit{\myblue{Detect whether this video contains  }}}\\
      \textbf{Event}            &Perception   & \multicolumn{2}{l}{\textit{\myblue{abnormal events or only normal events, and then give a description of the detected events with details, }}}\\
      \cline{2-2}
      \textbf{Description}      &Video        & \multicolumn{2}{l}{\textit{\myblue{especially environment, human looking and action.}}}\\
      \textbf{}                 &Synthesis    & \multicolumn{2}{l}{~~Truth: An \darkGreen{arrest} operation occurred in a public area. \myred{A male police officer} in a brown uniform confronted }\\
      \cline{1-2}
      \textbf{}                 &       & \multicolumn{2}{l}{\darkGreen{a man in a white T-shirt}, who was holding a bag, while \myred{a woman in white clothes} observed. The officer \darkGreen{} }\\
      \textbf{Anomaly}          &\multirow{2}{*}{Video}       & \multicolumn{2}{l}{\darkGreen{subdued the man, forcing him to the ground}. Another individual intervened, \myred{pointing at the suspect}. Later, }\\
      \textbf{Description}      &\multirow{2}{*}{Synthesis}       & \multicolumn{2}{l}{two black police officers joined,  with one putting on gloves to assist in lifting  the man and pushing him     }\\
      \textbf{}                 &      & \multicolumn{2}{l}{against a wall. \darkGreen{The man showed resistance}, but additional officers helped to handcuff him. Eventually, the  }\\
      \textbf{}                 &      & \multicolumn{2}{l}{ handcuffed man was placed on  a \darkGreen{stretcher} and transported by the police.}\\
      \Xhline{1.0pt}

    \end{tabular}
  }}
  \caption{An elaboration on UCVL's QA design. \myblue{Blue} marks the frozen question sentences across all videos, 
            while the question sentences for MCQs are automatically generated by the LLM. The truth for description questions are the summary of UCA‘s annotations.
            The \myred{red} phrases refer to the distractor options of the MCQs, while the \darkGreen{green} phrases refer to the correct option.
            }
    \label{table:QADesk}
\end{table*}

 \subsection{Multimodal Large Language Models}
 The capabilities of MLLMs have been extended to handle multi-image sequences and videos without altering the structure used for single-image inputs\cite{llava-onevision,internvl}.
While MLLMs generally adopt a universal architecture that includes a ViT encoder, an LLM decoder, and a connector that maps visual tokens to text embeddings, they differ in terms of training procedures and data.
 LLaVA-OneVision\cite{llava-onevision} and InternVL2\cite{internvl}  obtain video capability through finetuning on broad range of open-source academic datasets, including single-image, multi-image, and video of various tasks. 
In contrast, Qwen2-VL\cite{qwen2VL} employs a three-stage training approach with distinct alignment processes, leveraging a proprietary data configuration.

 \subsection{MLLM Benchmark}
Quantitative evaluation of MLLMs relies on benchmarks designed to assess the model’s capabilities across diverse topics and aspects. Each benchmark is typically assigned to a specific category, such as General QA, Document/Chart, Math, Code, OCR, Grounding, and others, with certain methods for scoring open-ended responses.  We compare 4 video benchmarks in Table \ref{table video large model benchs}. To assess models' responses, image-text benchmark VQAv2\cite{VQAv2} simplifies scoring by using standard and concise answers such as yes/no, 1/5 and kite/surfboard. 
  Meanwhile, video-MME\cite{video-mme} and many other benchmarks predominantly rely on multiple-choice questions (MCQs) for evaluation. 
 Additionally, MM-Vet\cite{mm-vet} leverages GPT for long-text comparison with few-shot instructions, which is also integrated in MMBench-Video\cite{mmbench-video}.

 Despite the abundance of image-text benchmarks, video benchmarks for MLLMs remain limited in number and have yet to include topics like anomaly analysis or the assessment of open-ended responses.

 \section{Method}

 \subsection{Data Collection}
 Our UCVL dataset builds upon the UCC\cite{UCFCrime} and UCA\cite{UCA} datasets.
 UCC includes 1,900 videos labeled as one of 13 anomalies or normal.
 UCA refines the data, segmenting each video into events with detailed annotations and timestamps.
 UCVL follows UCA's split and removes corrupted samples, obtaining a total of 1,699 videos. 
 We employ Qwen2-72B\cite{qwen2VL} and GPT-4o to summarize these descriptions and generate multitask QAs for MLLMs based on the combination of both datasets. The statistics of the resulting dataset are shown in Table \ref{tab:statistics}.
 We also develop efficient metrics for open-ended evaluation. The details of our work are as follows.

 \subsection{Video Summarization}
 UCA segments the video and provides straightforward and independent annotations for each segment, which diminishes the coherence and logical flow of the narrative.
  Moreover, redundant descriptions often overshadow key events and actions with excessive details, making event analysis challenging. 
  Most critically, UCA fails to explicitly declare or classify crimes that could be supplemented by UCC. An example of UCA is shown in Fig. \ref{fig:pipeline}.

 \begin{table}[t]
  \centering
  \renewcommand{\arraystretch}{1.2} 
  \setlength{\tabcolsep}{3pt} 
  \begin{tabularx}{\columnwidth}{@{}XllllX@{}}
  \toprule
  \textbf{Benchmark}  & \textbf{MVBench} &\textbf{Video-MME}& \textbf{MMB-Video}& \textbf{UCVL} \\
  \midrule
  \textbf{Videos}          & 200& 900& 600& 1699 \\
  \textbf{QA Pairs}        & 4000& 2700& 2000& 16990 \\
  \textbf{Content}          & Normal& Normal& Normal& Anomaly \\
  \textbf{QA Forms}        & MCQ& MCQ& open-ended& MCQ\&open-ended \\
  \textbf{Generation}   & LLM & Human& Human& LLM  \\
  \textbf{Evaluation}      & Matching& Matching& LLM& Matching\&LLM \\
  \bottomrule
  \end{tabularx}
  \caption{A comparison of video benchmarks for MLLMs. `Normal' content indicates that the dataset covers a wide range of topics without a specific focus. `Matching' is short for fuzzy pattern matching.}
    \label{table video large model benchs}
\end{table}

 Therefore, we first combine UCC's crime labels and UCA's descriptions and timestamps to collaboratively generate a high-quality summary of the video. 
 This summarization highlights anomalies, 
describes the video concisely and logically, emphasizes key evidence, and identifies the start/end times of the anomalous events.
 We employ Qwen2-72B to perform the process via few-shot prompting. The detailed prompts and examples are provided in Appendix B.
 To validate the feasibility of this approach, we compare the generated results with human annotations in Appendix A. While human annotators provide more detailed information about the characters, the LLM's summaries clearly and accurately convey the basic features and actions of the events.

 
      

 \subsection{Question-Answer Generation}
 
 To evaluate MLLMs' capabilities in perceiving and analyzing anomalies,
 we designed six types of QAs: anomaly detection (True or False, TF), anomaly classification (AC), anomaly temporal grounding (TG), multiple-choice questions  (MCQ), event description (ED), and anomaly description (AD). A sample of six QA types for a video is shown in Table \ref{table:QADesk}.

 TF asks whether the video contains evidence of anomalies.
 AC provides fourteen categories for the model to select the most appropriate.
 TG uses LLaVA's time instruction to provide the model with the timestamps of each frame and then asks it to identify the start/end times of the main event in the video.
 The answers to the above three questions are the labels from UCC or the begin/end time from our summaries.

 To assess a model's recognition of crime concepts, we set ED and AD as an ablation study.
 ED prompts the model to describe the video's main event in detail and assess whether it is abnormal, while AD informs the model of the anomaly category and then asks the same question as ED.
 The summaries serve as the ground truth for both questions, which are concise and meaningful for depicting the anomaly event.
   \begin{table}[t]
\centering
\setlength{\tabcolsep}{4mm}
\begin{tabular}{@{}rllll@{}}
\toprule
\textbf{Item} & \textbf{Train} & \textbf{Val} & \textbf{Test} \\ 
\midrule
Videos         & 1030 &369 & 300  \\
Video length &50.2h & 18h &20h \\
Max length   &107min &130min &94min  \\
Min length   &3.5s & 8.5s &4.6s  \\
QA pairs       &10300 &3690 &3000  \\
Words/Summary  &75  &65 &70 \\

\bottomrule
  \end{tabular}
   \caption{The statistics of our UCVL dataset.}
  \label{tab:statistics}
\end{table}

 Following the mainstream of MLLM benchmarking, we generate five MCQs for each video, focusing on four key aspects related to case investigation: causal reasoning, object detail recognition, action capturing and counting. See Fig. \ref{fig:pipeline} for prompt details.
 We employ Qwen2-72B to generate both questions and answers based on the facts from video summaries.

 \subsection{Evaluation Metrics}
 To evaluate the model's responses, we apply different metric designed to each question type.
 
 \textbf{Objective Metrics.} 
 For TF and MCQ, we directly calculate the accuracy. 
 For TG, we use Intersection over Union (IoU) to measure the degree of overlap between 2 time segments.
 For AC, since the anomaly facts may relate to multiple categories, we evaluate the model using a top-3 accuracy metric.

 \textbf{Subjective Metric.} 
 For ED and AD, the evaluation methods are identical.
 In line with common practices of MLLM benchmarks, we use GPT-4o for comparing open-ended text responses.
 We go beyond other works by not only providing scoring examples but also incorporating detailed scoring guidelines, ensuring stable and precise evaluation. The details are shown in Appendix B.

All task scores are normalized to a scale of 0-100, with weights assigned based on the importance of each task. Consequently, the weighted sum of scores (Total) for each model is calculated as follows:
\begin{align*}
\text{Total} = &0.15 \cdot S_{TF} + 0.1 \cdot S_{AC} + 0.15 \cdot S_{ED} + \\
&0.15 \cdot S_{AD} + 0.2 \cdot S_{TG} + 0.05 \cdot 5 \cdot S_{MCQ} 
\end{align*}
where  $S_{*}$ denotes the score of a task. 
 
 \section{Experiment}

 \subsection{Experimental Setup}
 We conduct evaluation on eight open-source video models, including InternVL2\cite{internvl}, Qwen2-VL\cite{qwen2VL}, and LLaVA-OneVision\cite{llava-onevision} series, with parameter sizes ranging from 0.5B to 40B. 
 For video sampling, we adopt a fixed-frame strategy at 8, 16, 32, and 64 frames. 
 The models infer one question at a time to prevent information leakage between questions.
 We complete the evaluation on the test set with a total of 3,000 questions. All experiments are conducted on NVIDIA H800 GPUs.

\begin{table}[htbp]
  \centering
  \renewcommand{\arraystretch}{1.2} 
  \setlength{\tabcolsep}{0.6mm} 
  \setlength{\fboxsep}{1pt}
  \begin{tabularx}{\columnwidth}{@{}X|c|ccccccccX@{}}
  \toprule
  \textbf{Model} &\textbf{MMB}& \textbf{TF} & \textbf{AC} & \textbf{MCQ} & \textbf{ED} & \textbf{AD} & \textbf{TG/\%} & \textbf{Total} \\ \midrule
  LLaVA-OV-0.5B            &52.1            & 82.7          & 32.3          & 56.5          & 30.9          & 18.3          & 17.0          & 40.5 \\ 
  InternVL2-1B             &65.4            & 56.7          & 30.0          & 46.1          & 21.2          & 25.0          & 2.4          & 30.5 \\ 
  Qwen2-VL-2B              &74.9            & 83.7          & 46.7          & 64.7          & 34.3          & 37.3          & 4.9          & 45.1 \\ 
  Qwen2-VL-7B              &83.0            & 81.0          & 50.0          & 69.8         & 42.9          & 53.6          & 19.6          & 52.0 \\
  LLaVA-OV-7B              &80.8            & 63.3          & 43.7          & 64.7          & 35.3          & 38.6          & 8.3          & 44.1 \\
  InternVL2-8B             & \fbox{81.7}            & 83.7          & 46.7          & 64.6          & 34.2          & 39.3          & 17.9          & 48.0 \\ 
  InternVL2-26B            &\underline{83.4}            & \fbox{86.3}          & \underline{55.0}          & 66.4          & \textbf{57.7} & \underline{65.1}          & \fbox{26.3}          & \underline{58.7} \\ 
  InternVL2-40B            &\textbf{86.8}   & \underline{89.3}          & \textbf{56.7} & \underline{71.7}          & \fbox{47.3}          & \textbf{72.9} & 13.7          & \fbox{57.8} \\
  \midrule  
  LLaVA-UCVL-0.5B          &46.2            & 74.3          & 31.0          & \fbox{71.3}         & 27.6          & 35.3          & \underline{42.7}          & 50.1 \\
  LLaVA-UCVL-7B            &80.7            & \textbf{91.3} & \fbox{54.0}          & \textbf{75.5} & \underline{49.9}          & \fbox{54.8}          & \textbf{50.5} & \textbf{63.8}  \\ 
  \bottomrule
  \end{tabularx}
   \caption{Evaluation results of 10 models on UCVLBench. MMB denotes MMBench which represents the general 
   performance. TF, AC, MCQ, ED, ED, AD, TG denote 6 Question-Answer types. Total is a weighted sum of the 6 scores. 
   \textbf{Bold} denotes the best score in this QA type, \underline{underline} the second best, and \fbox{box} the third.}
\label{table_32frame_eval}
 \end{table}

 \subsection{Evaluation Results}
We present the results for each task at 32 frames, as shown in the Table \ref{table_32frame_eval}.
We list eight models in order of parameter sizes and collected their performance metrics on MMBench as a reference. 

Within the same series, the results generally indicate a correlation between model performance and parameter sizes. 
However, an exception is observed between the 26B and 40B models of InternVL2, due to 40B's shortage in ED and TG. 

Across model series, results reveal the models' blindness in anomaly despite their general capabilities.
LLaVA-OV-0.5B outperforms InternVL2-1B with fewer parameters and lower general performance. Similarly, Qwen-VL-2B outperforms LLaVA-OV-7B.
This trend is particularly evident in models smaller than 8B, indicating they may possess greater potential for domain specialization than their larger counterparts.

\begin{figure}[ht]
  \begin{flushleft}
  \includegraphics[width=0.9\linewidth]{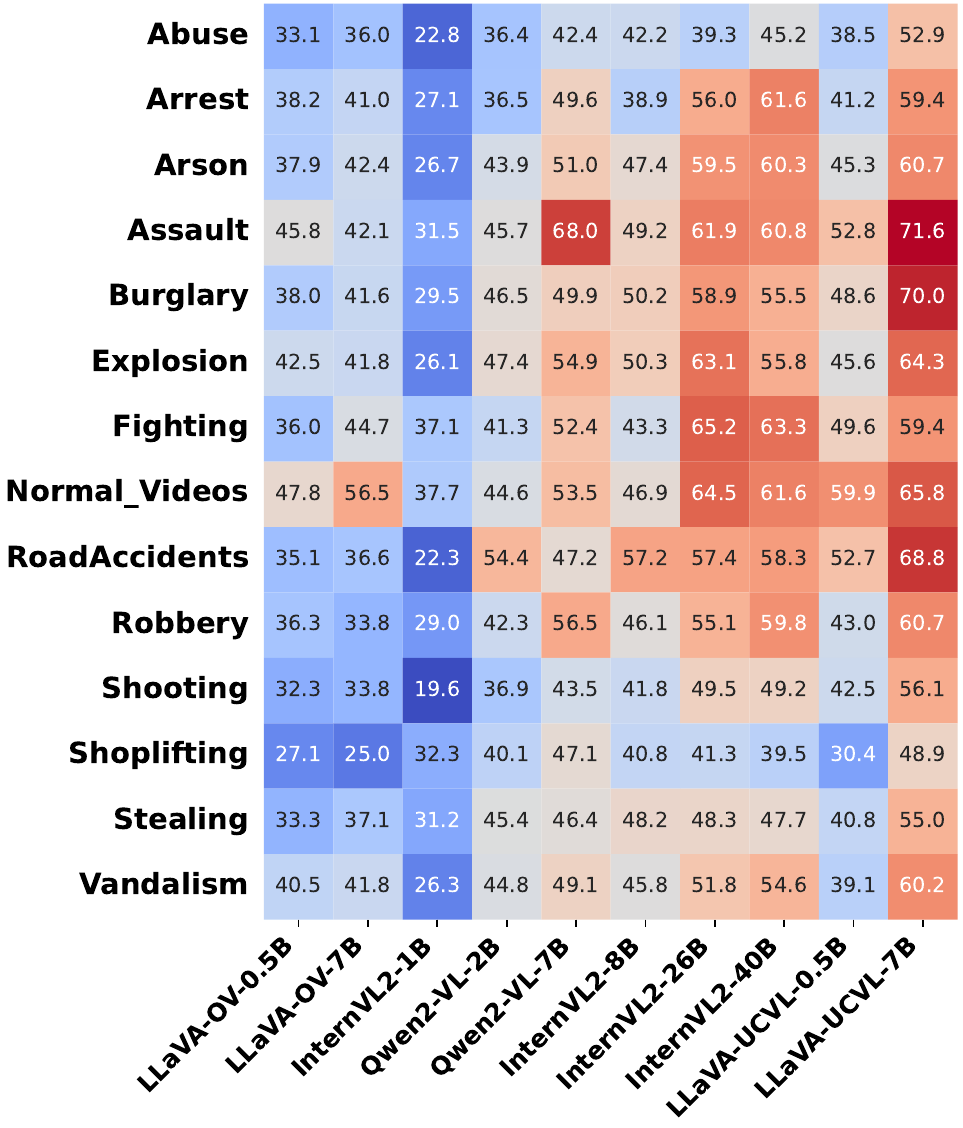}
  \caption{Heatmap of models' performance across 14 crime categories. }
  \label{fig:heatmap}
  \end{flushleft}
\end{figure}

Fig. \ref{fig:heatmap} illustrates the models' performance on 14 crimes. Overall, models score notably higher on crimes with prominent actions such as assault, fighting, road accident and explosion. Furthermore, the models perform better on normal videos than on crime-related ones, emphasizing their limited ability to perceive criminal activities.


\subsection{Finetuning Results}
For finetuning on the training set, we use LLaVA-OneVision's 0.5B and 7B models as baselines. 
To improve training efficiency, we convert MCQs into multi-round conversations and merge ED and AD into a single video description task. The other tasks are retained as single-round conversations.
We finetune 7B model at 32 frames on four NVIDIA H800 GPUs for an hour, and 0.5B model on two NVIDIA H800 GPUs for half an hour. The learning rates are 3e-6 for the 7B model and 1e-6 for the 0.5B model, with batch sizes of 32 and 16, respectively.

We name the finetuned models as LLaVA-UCVL. 
The 7B model achieves substantial improvements across all tasks, outperforming InternVL2-40B.
This highlights the benefits of targeted finetuning for anomaly analysis tasks. A qualitative result is shown in Fig. \ref{fig:fig1}.
Meanwhile, the 0.5B model shows notable improvements in MCQ, AD, TG and Total, surpassing InternVL2-8B,
but exhibits slight declines in TF, AC, and ED tasks, which rely on anomaly detection capabilities. A decrease in overall performance is also reflected in the MMBench score.
We attribute this to the small parameter size, which leads to overfitting. A potential solution is to merge this dataset into LLaVA's fine-tuning phase with a certain proportion, which we plan to explore in future work.


 \subsection{Ablation Study}

 \textbf{Ablation on number of sampled frames.}
As shown in Table \ref{table:nframes_ablation},
generally, the models' performances increase as frame number increases.
However, when it increases to 64, smaller models begin to see a decline in performance, while larger models show a slight improvement. The increase reflects better capture of the action process, while the decline suggests limited capability to handle longer inputs. We provide the frame numbers used in the models' training process to explain this. It also indicates that larger models can achieve better performance even with more frames than during training.

From the perspective of the benchmark's reliability, the observed trends across different frame numbers appear consistent and reasonable, and the finetuned models closely resemble the original trends,
 which shows that UCVL accurately reflects the models' performance.

 \begin{table}[htbp]
  \centering
  \renewcommand{\arraystretch}{1.2} 
  \setlength{\tabcolsep}{2pt} 
  \begin{tabularx}{\columnwidth}{>{\raggedright\arraybackslash}p{2.5cm}>{\centering\arraybackslash}X*{4}{>{\centering\arraybackslash}X}} 
  \toprule
  {}  &\multirow{2}{*}{Training} & \multicolumn{4}{c}{UCVL Score on Nframes} \\ 
  \cmidrule(lr){3-6}
  \textbf{Model}& Nframes &\textbf{8} & \textbf{16} & \textbf{32} & \textbf{64} \\ \midrule

   LLaVA-OV-0.5B    &32   & 34.6  & 33.5  & \textbf{40.5}  & 32.2 \\
   InternVL2-1B     &8-32 & \textbf{32.1}  & 30.3  & 30.4  & 28.9 \\
   Qwen2-VL-2B      &/     & 38.7  & 40.4  & \textbf{45.1}  & 43.0 \\
   Qwen2-VL-7B      &/     & 44.8  & 48.2  & \textbf{52.0}  & 48.5 \\
   LLaVA-OV-7B      &32    & 38.3  & 41.0  & \textbf{44.1}  & 42.2 \\
   InternVL2-8B     &8-32  & 44.6  & 47.6  & 48.0  & \textbf{49.1} \\
   InternVL2-26B    &8-32  & 49.6  & 52.8  & 58.7  & \textbf{60.3} \\
   InternVL2-40B    &8-32  & 50.7  & 52.2  & 57.8  & \textbf{59.4} \\ \hline
   LLaVA-UCVL-0.5B  &32    & 44.5  & 43.7  & \textbf{50.1}  & 41.8 \\
   LLaVA-UCVL-7B    &32    & 61.4  & 63.2  & 63.8  & \textbf{64.2} \\
   \bottomrule
 \end{tabularx}
    \caption{Evaluation results of ablation study on the number of uniformly sampled frames from a video. \textbf{Bold} denotes the best score of this model.}
    \label{table:nframes_ablation}
 \end{table}

 \textbf{Ablation on ED and AD.}
Table \ref{table_32frame_eval} shows models' performance on ED and AD under the sampling strategy at 32 frames. 
The slight difference between the two questions is that we inform the model of the anomaly's category at the beginning of the AD prompt: \textit{We believe that an event of [label] happens in this video.}
Typically, models achieve higher scores on AD than on ED, except for LLaVA-OV-0.5B due to its weak instruction-following ability.
This performance gap between AD and ED highlights a deficiency in the models' ability to perceive anomaly events. Nonetheless, finetuning mitigates this issue, as LLaVA-UCVL-7B achieves a score of 49.9 on ED and 54.8 on AD, narrowing the gap and surpassing InternVL2-40B's score of 47.7 on ED.

\bigskip

 \section{Conclusion}
In this paper, we present UCVL, the first benchmark designed to evaluate the capabilities of MLLMs in video anomaly analysis, offering a diverse range of tasks and question types. To assess the models' open-ended text responses, we developed a detailed scoring system powered by GPT-4o.
The evaluation results not only highlight the models' potential in this domain but also validate the robustness of UCVL. Moreover, our models finetuned on UCVL's training set, effectively harness this potential and demonstrate significant improvements. 
For future research, this work paves the way for applying large models to video anomaly analysis.

\bibliographystyle{IEEEbib}
\bibliography{main}

\end{document}